\pdfoutput=1

\documentclass[12pt]{iopart}

\usepackage{graphicx}
\usepackage{subfigure}
\usepackage{cite}
\usepackage[export]{adjustbox}
\usepackage{xcolor}
\usepackage{hyperref}
\usepackage{multirow}

\begin{document}

\title[Author guidelines for IOP Publishing journals in  \LaTeXe]{Augmented reality navigation system for visual prosthesis}

\author{Melani Sanchez-Garcia*, Alejandro Perez-Yus, Ruben Martinez-Cantin, Jose J. Guerrero}


\address{Instituto de Investigación en Ingeniería de Aragón, (I3A). Universidad de Zaragoza, Spain}
\ead{mesangar@unizar.es}
%\vspace{10pt}
%\begin{indented}
%\item[]August 2017
%\end{indented}

\begin{abstract} \\

The visual functions of visual prostheses such as field of view, resolution and dynamic range, seriously restrict the person’s ability to navigate in unknown environments. Implanted patients still require constant assistance for navigating from one location to another. Hence, there is a need for a system that is able to assist them safely during their journey. In this work, we propose an augmented reality navigation system for visual prosthesis that incorporates a software of reactive navigation and path planning which guides the subject through convenient, obstacle-free route. It consists on four steps: locating the subject on a map, planning the subject trajectory, showing it to the subject and re-planning without obstacles. We have also designed a simulated prosthetic vision environment which allows us to systematically study navigation performance. Twelve subjects participated in the experiment. Subjects were guided by the augmented reality navigation system and their instruction was to navigate through different environments until they reached two goals, cross the door and find an object (bin), as fast and accurately as possible. Results show how our augmented navigation system help navigation performance by reducing the time and distance to reach the goals, even significantly reducing the number of obstacles collisions, compared to other baseline methods.

% On the other hand, very few studies are performed on the navigation capabilities that could be restored by these implants, probably for practical as well as safety reasons. 

%\url{https://reader.elsevier.com/reader/sd/pii/S0042698907001071?token=0FD4052AC969063F328393E48C4577762080D779415018006EC7DE918C228E63B0115285BBC938F58723FEB2A7895B6D}

\end{abstract}

\vspace{2pc}

\section{Introduction}

\begin{figure}[t]
\centering
\includegraphics[width=1\textwidth]{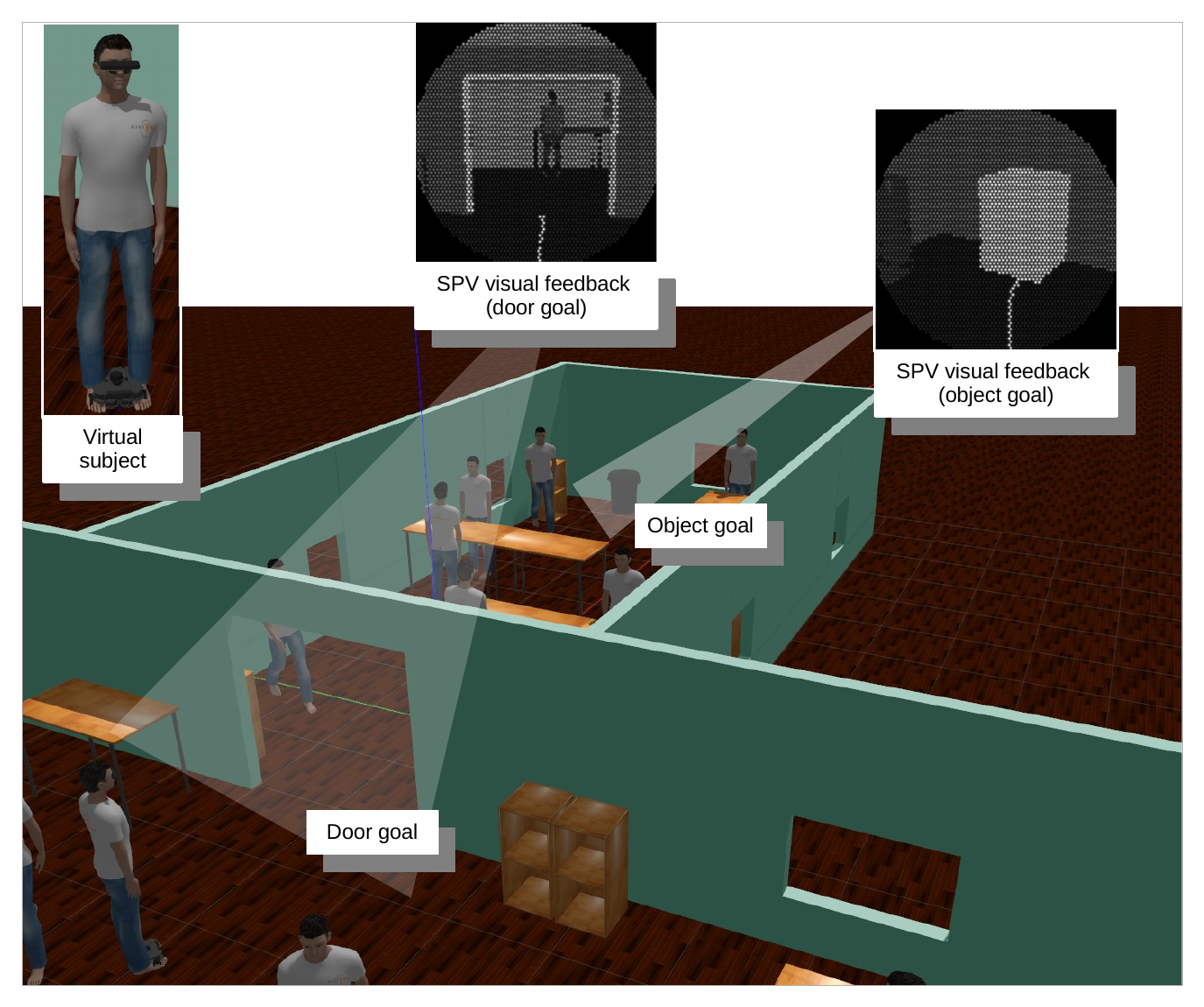}
\caption{\textbf{Overview.} Our simulated prosthetic vision system enhance visual feedback for the user as an aid in navigation tasks. We introduce an augmented reality navigation system for visual prosthesis (\textit{RoboticG}). This system consist in a path planning algorithm for routing the subject through an obstacle free and optimal path (augmented path) and a plane extraction of the goals (augmented goals) which are highlighted with a higher brightness level of the phosphenes.}
\label{fig1}
\end{figure}

% Realidad aumentada: añadimos cosas que no están ahí al usuario para darle información extra --> AR

Technological integration of Artificial Intelligence (AI) in the field of prosthetics and assistive technology has become a helpful tool for people with disabilities. Application of AI and robotics technology has a huge potential in achieving independent mobility and enhances the quality of life in persons with disabilities \cite{nayak2020application,buhler1995autonomous,park2007robotic,uehara2010mobile,morrison2017imagining}.

Augmented reality (AR) is showing day by day that has a place in our daily lives and can become a useful tool. Researchers have placed this new horizon to produce technology that is destined to change the way we solve problems and interact with the world. This ranges from work, education, medicine, and enhancing the independence of people with visual disabilities such as Retinitis Pigmentosa (RP) and Age-related Macular Degeneration (AMD) \cite{sanchez2020semantic,denis2014simulated,dramas2010artificial,mace2011brain}.

RP and AMD are the two most prevalent retinal degenerative diseases. They affect millions of individuals worldwide and cause permanent blindness due to a gradual loss of photoreceptor \cite{curcio2000spare,busskamp2010genetic,yue2016retinal}. Although current remedies can slow the progression of vision loss, there are no permanent cures for these retinal diseases \cite{richer2004double,heier2012intravitreal}. Retinal prostheses have turned out to a promising technology to improve vision in patients with RP and AMD. Visual prostheses can partially restore vision, bypassing damaged photoreceptors and electrically stimulating the surviving retinal cells, such as the retinal ganglion cells. For the blind, a retinal prosthesis can be life-changing, partially restoring sight, improving mobility and even helping to recognize some objects. The Argus II epiretinal prosthesis (Second Sight Medical Products Inc., Sylmar, CA) combines a miniature eye implant with a patient-worn camera and a processor to transform how the patient experience the world \cite{luo2016argus}. The camera acquires images from the real-world that are transmitted to a portable visual processing unit linked to the camera. The processed information is sent to the retina via electrical impulses in the implant by an electrode array. The stimulation can activate a group of neurons in a small localized area of the retina. Patients implanted with an array of electrodes in the visual system report the perception of reproducible white spots in the visual fields (called phosphenes) after stimulation. The brain interprets patterns of phosphenes in the restricted area as visual information. To date, the results of implanted subjects have shown to be promising, and all subjects have demonstrated improved function using the implanted prosthesis.

There are still physiological and technological limitations of the information received by implanted patients. Spatial resolution of prosthetic implants is limited by several factors, including electrode density, size, number and pitch, electrode contact, and visual encoding \cite{bloch2019advances}. Up to 600 or 1,000 pixels are required for restoration of useful vision; however, the majority of prostheses use 100 or fewer stimulating electrodes \cite{weiland2011retinal}. Besides, current systems provides a field of view of approximately $18^\circ$ x $11^\circ$ in the retinal area, which correspond to the field of view covered by the electrode implant on the retina.

Because of the small number of implanted patients, it is difficult to improve prosthesis design through extensive clinical trials. To systematically progress on the design of retinal prostheses it is possible to rely on Simulated Prosthetic Vision systems (SPV). SPV opens the opportunity to evaluate potential and forthcoming functionality in early stages of design with not implanted subjects.

The SPV system usually consists in a computer screen for the presentation of static or dynamic phosphene images or in a camera mounted on a virtual-reality headset \cite{li2006simulated,maeder2004mobility,mccarthy2012time,mccarthy2011ground,feng2013enhancing,vergnieux2017simplification,sanchez2020semantic}. Participants perform various tasks perceiving a set of phosphenes mimicking the percepts elicited by a retinal prosthesis, while wearing the head-mounted display. For example, Fornos et al. \cite{fornos2008simulation} used SPV to study how the restrictions of the amount of visual information provided would affect performance on simple pointing and manipulation tasks. Hayes et al. \cite{hayes2003visually} designed a set of tasks to assess performance of object recognition and manipulation and reading using different sizes of electrode array. In a posterior study, Dagnelie et al. \cite{dagnelie2007real} explored minimal visual resolution requirements of a simulated retinal electrode array for mobility in real and virtual environments, experienced by normally sighted subjects in video headsets. Sanchez-García et al. \cite{sanchez2020influence} evaluated the influence of field of view with respect to spatial resolution in visual prostheses using a virtual-reality system for more realistic SPV environments using panoramic scenes. However, they mainly focused on reading and object recognition.

More and more SPV navigation studies are being carried out \cite{vergnieux2012spatial,vergnieux2017simplification,wang2008virtual}. Navigation is an important component of the self-reliance of the blind. The visual functions of patients are restricted enough to seriously affect the person's ability to navigate unknown environments. Some researchers have focused on mobility and obstacle avoidance. Cha et al. \cite{cha1992mobility} investigated the feasibility of achieving visually-guided mobility without extra information. They showed that 625 phosphenes and a camera field of view of 30$^{\circ}$ were required to reach acceptable performances. More recent studies also focused on obstacle avoidance in a corridor \cite{barnes2011investigating,parikh2013performance}. Other works showed that highlighting obstacles \cite{mccarthy2014mobility} or planar surfaces \cite{mccarthy2010surface} improve the preferred walking speed \cite{clark1986efficiency}. Dagnelie et al. \cite{dagnelie2007real} explored minimal visual resolution requirements of a simulated retinal electrode array for mobility in real and virtual environments with a high contrast between the ground and the walls. Rheede et al. \cite{van2010simulating} also used a virtual environment to check if subjects were able to follow instructions and walk through a predetermined path. The results of Dagnelie et al. \cite{dagnelie2007real} and Rheede et al. \cite{van2010simulating} also demonstrated that the comprehension of the environment was not necessary for the subjects to follow a predetermined path. Successful localization is based on the perception of specific signals from the environment (landmarks), but also on the selection of an appropriate path \cite{meilinger2008network}. This is precisely the trend that has been followed in Robotics in recent years \cite{bagnell2010learning,arkin1990autonomous,piaggio2001autonomous,silver2012active,cheng2018autonomous}.

% Navigation and obstacle avoidance are difficult to perform with low resolution implants, which points out the need to highlight pertinent information within the surroundings. 

%https://hal.archives-ouvertes.fr/hal-01692744/document

The problem of navigation in Robotics has been extensively studied. We can define autonomous navigation as a set of methodologies that make possible to move a robot safely through the environment. Autonomous navigation and obstacle avoidance have been widely study in mobile robotics \cite{lavalle2006planning,karaman2011sampling,hoy2015algorithms,paden2016survey}. More concretely, the aim of navigation systems is to search an optimal or quasi-optimal path from the start point to the goal point with obstacle avoidance competence. The autonomous navigation system usually includes various tasks such as: planning, perception and control. Planning in mobile robots generally consists of establishing the mission, the route and the avoidance of obstacles, sometimes in the presence of uncertainty \cite{everett2018motion,bai2015intention}. In the case of robotics, autonomous navigation culminates by sending orders to the robot so that it moves in the calculated direction. In our case of people assistance, we look for a way to communicate the information obtained through a visual prosthesis. 

% For instance, several researches have been conducted in order to avoid collisions and navigate autonomously using Simultaneous localization and mapping (SLAM), Robot Operating System (ROS) and simulation environments such as Gazebo \cite{ocando2017autonomous,ibanez2017implementation,lee20172d,santos2013evaluation}.

% file:///C:/Users/melan/AppData/Local/Temp/(14-18)Autonomous%20Navigation%20with%20Collision%20Avoidance%20using%20ROS-format.pdf

In this work, we take advantage of the robotic algorithms, adapt them to human navigation, and include them in SPV using a virtual environment. Specifically, we propose an augmented reality navigation system for visual prostheses. This method consists on four tasks: locating the subject on a map, planning the subject trajectory, showing it to the subject and re-planning without obstacles. We also design a virtual environment which allows us to systematically study navigation performance based on prosthetic vision. The main issue is to determine whether the use of a smart prosthesis based on a robotic navigation system help in navigation tasks. We evaluate and compare the proposed guidance system with baseline methods through a SPV experiment, which is a standard procedure for non-invasive evaluation using normal vision subjects. The experiments included one task: finding and reaching a large object while avoiding obstacles in wildly crowded scenarios.

\begin{figure}
\centering
\subfigure[]{\includegraphics[width = 2in]{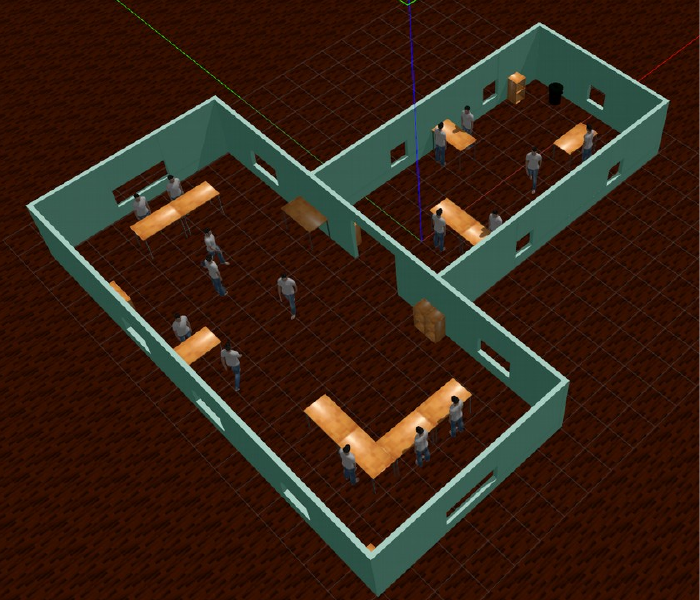}\label{fig2a}}
\subfigure[]{\includegraphics[width = 2in]{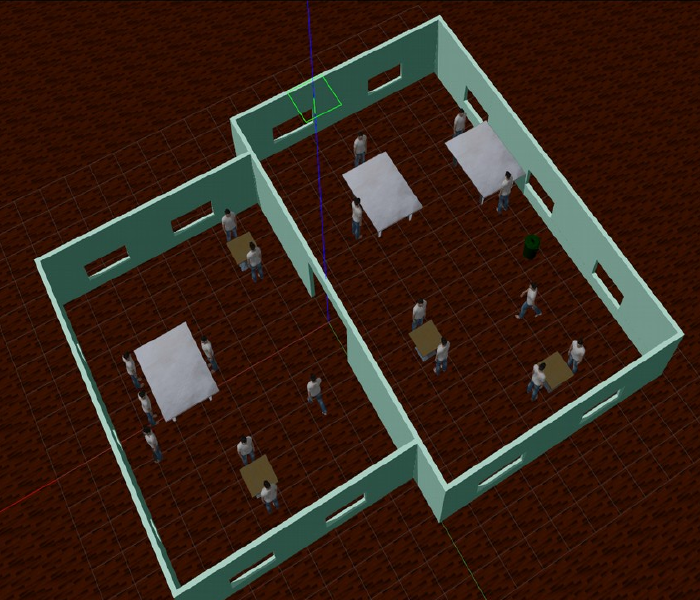}\label{fig2b}}
\subfigure[]{\includegraphics[width = 2in]{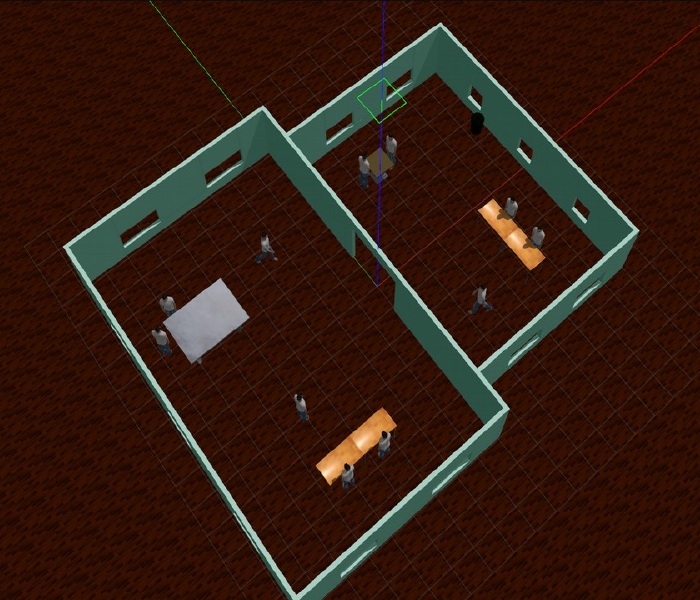}\label{fig2c}}
\caption{\textbf{ Virtual environments used in the experiments.} \subref{fig2a} Environment 1, \subref{fig2b} Environment 2 and \subref{fig2c} Environment 3. Both, Environment 2 and Environment 3 have the same map but vary in the number and distribution of obstacles, being the Environment 2 more complicated than Environment 3, but less than Environment 1.}
\label{fig2}
\end{figure}

% Methods
\section{Methods}

% Subjects
\subsection{Subjects}
Twelve subjects with normal vision volunteered for the formal experiment. The subjects (two females and ten males) were between 22 and 35 years old. Every subject used a computer daily (video games). 

% Ethical statement
\subsubsection{Ethical statement}

The research process was conducted according to the ethical recommendations of the Declaration of Helsinki. The research protocol used for this study is non-invasive, purely observational, with absolutely no-risk for any participant. There was no personal data collection or treatment and all subjects were volunteers. Subjects gave their informed written consent after explanation of the purpose of the study and possible consequences. The consent allowed the abandonment of the study at any time. All data were analyzed anonymously. The experiment was approved by the Aragon Autonomous Community Research Ethics Committee (CEICA, see Ethical Statement for additional details).

% Stimuli
\subsection{Augmented Reality System}

\begin{figure}
\centering
\subfigure[]{\includegraphics[width = 2in]{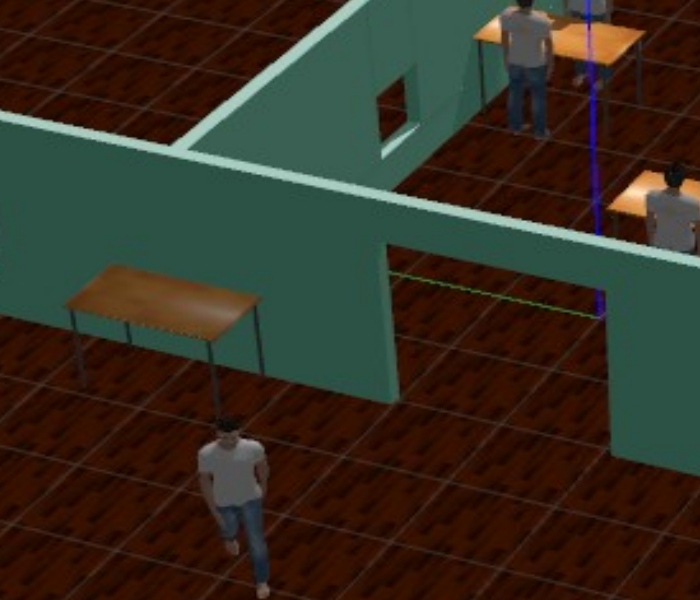}\label{fig3a}}
\subfigure[]{\includegraphics[width=2in,height=1.73in]{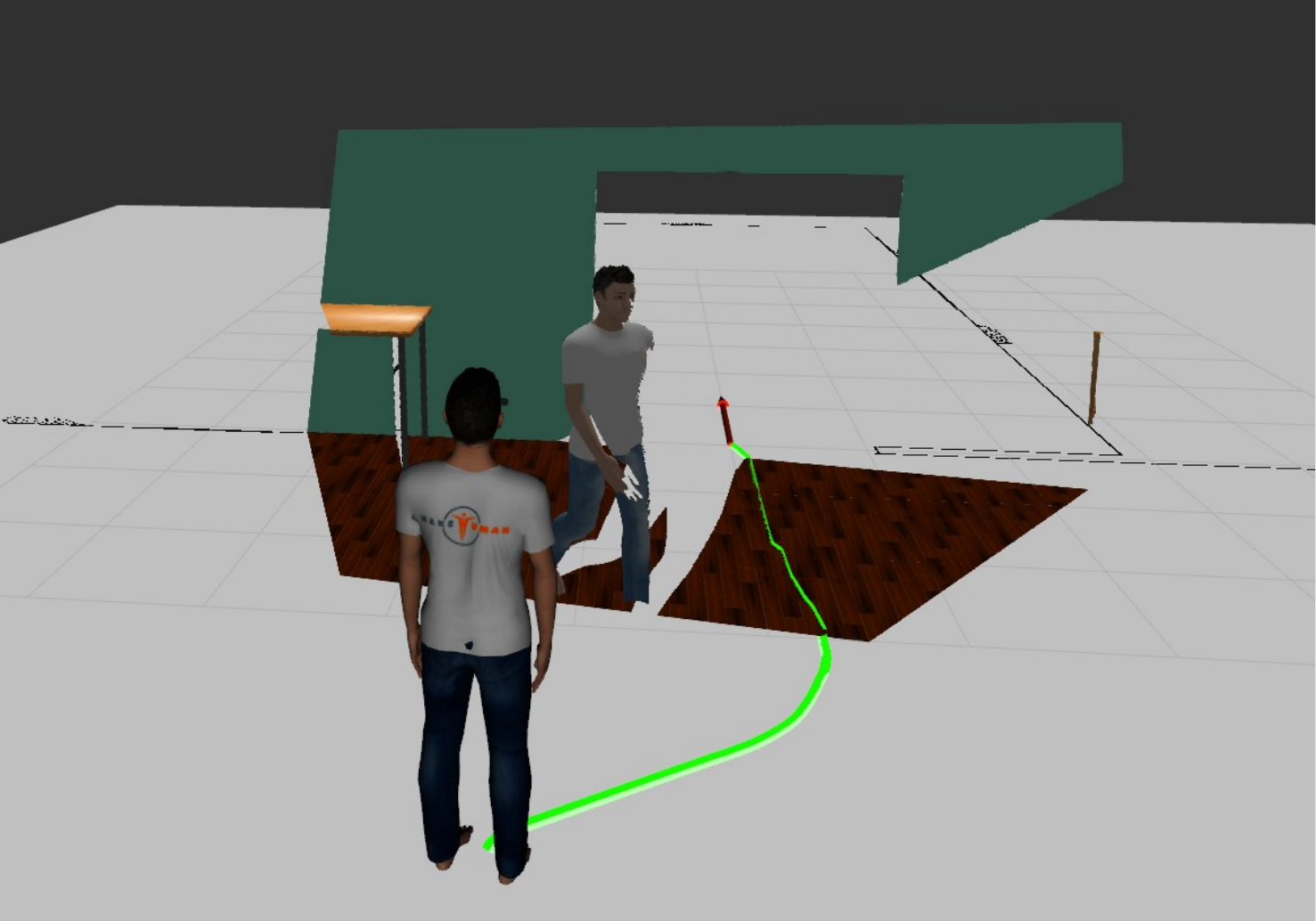}\label{fig3b}}
\subfigure[]{\includegraphics[width = 2in]{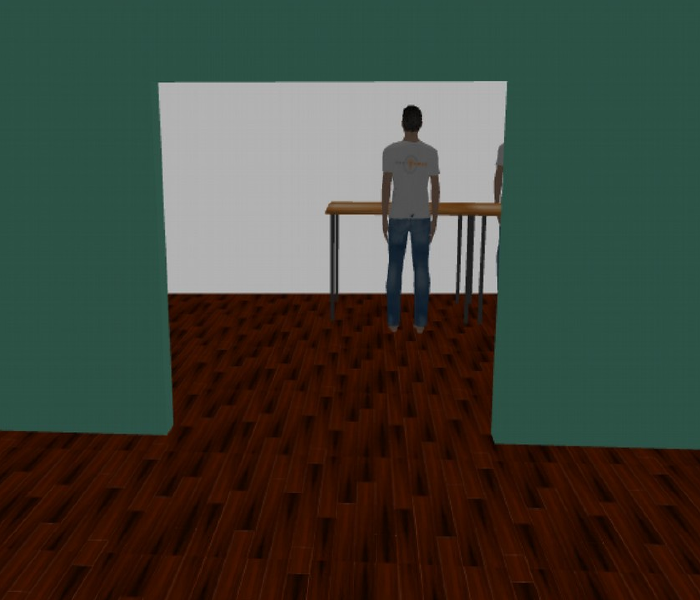}\label{fig3c}}
\subfigure[]{\includegraphics[width=2in]{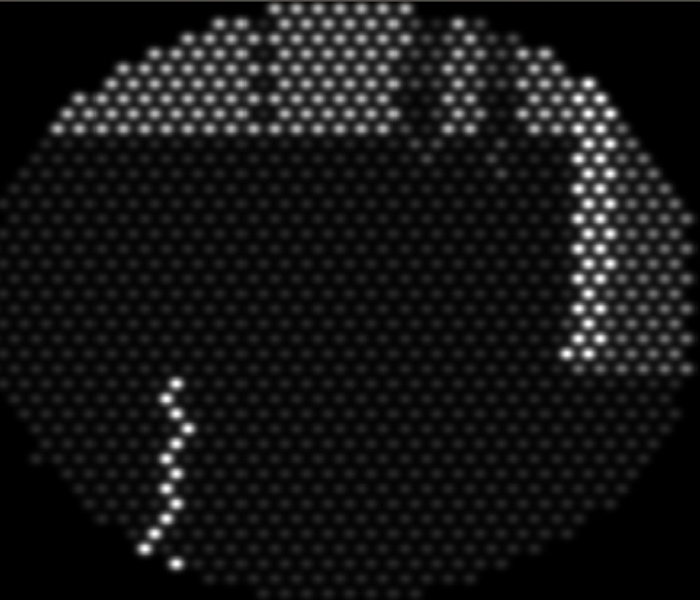}\label{fig3d}}
\subfigure[]{\includegraphics[width = 2in]{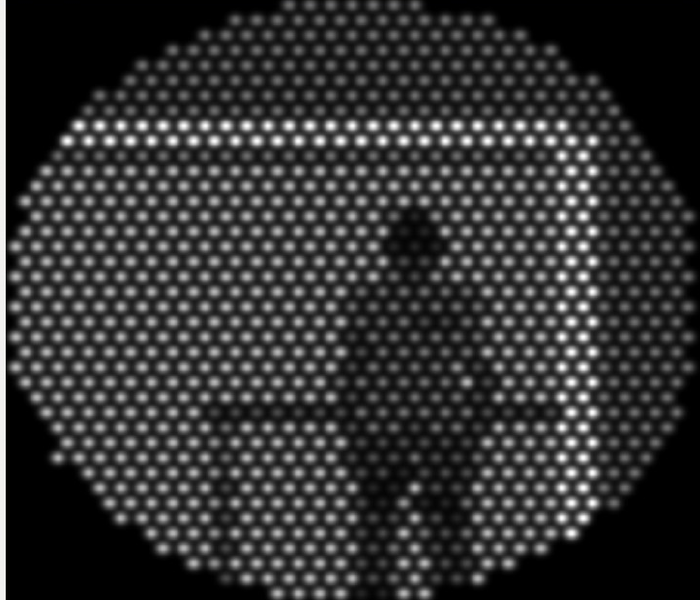}\label{fig3e}}
\subfigure[]{\includegraphics[width = 2in]{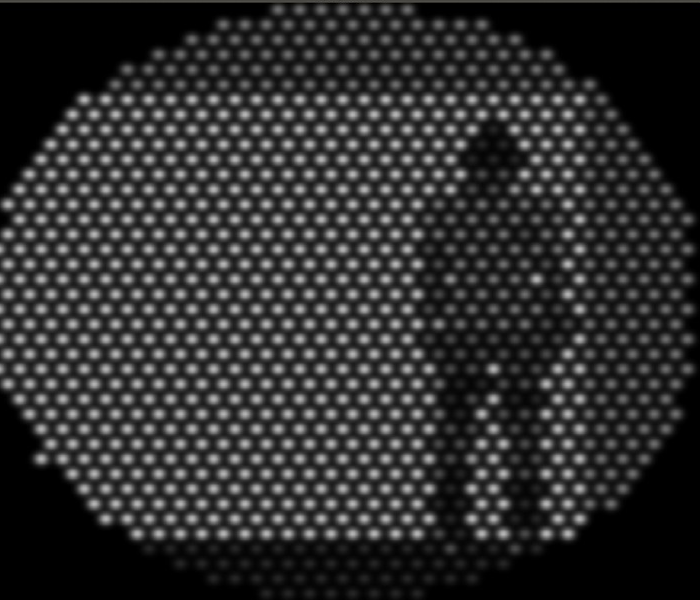}\label{fig3f}}
\caption{\textbf{Guidance methods with door goal.} In this example, the subject's goal is to cross the door. \subref{fig3a} Simulation environment showing the door goal. \subref{fig3b} The robotic system obtains an optimized trajectory to the goal avoiding obstacles (green line). \subref{fig3c} Image captured by RGB camera showing the door goal. \subref{fig3d} Image generated by the SPV with the RoboticG method. \subref{fig3e} Image generated by the SPV with the PerceptualG method. \subref{fig3f} Image generated by the SPV with the DirectG method. }
\label{fig3}
\end{figure}

\begin{figure}
\centering
\subfigure[]{\includegraphics[width = 2in]{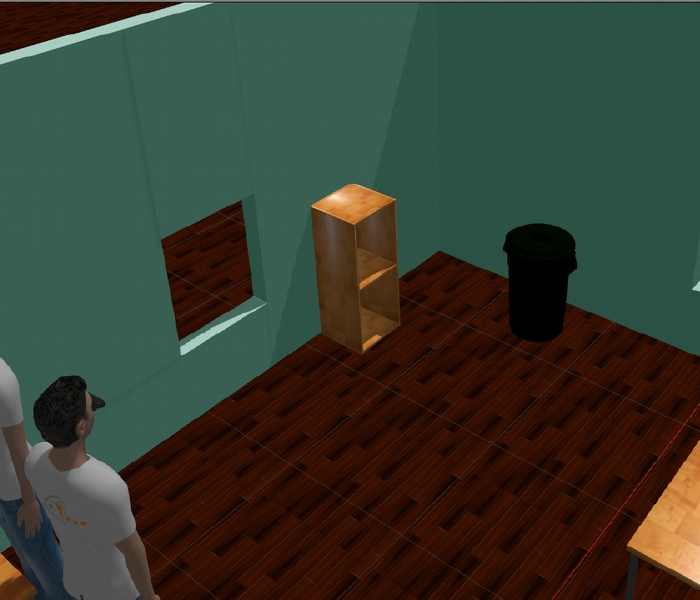}\label{fig4a}}
\subfigure[]{\includegraphics[width=2in,height=1.73in]{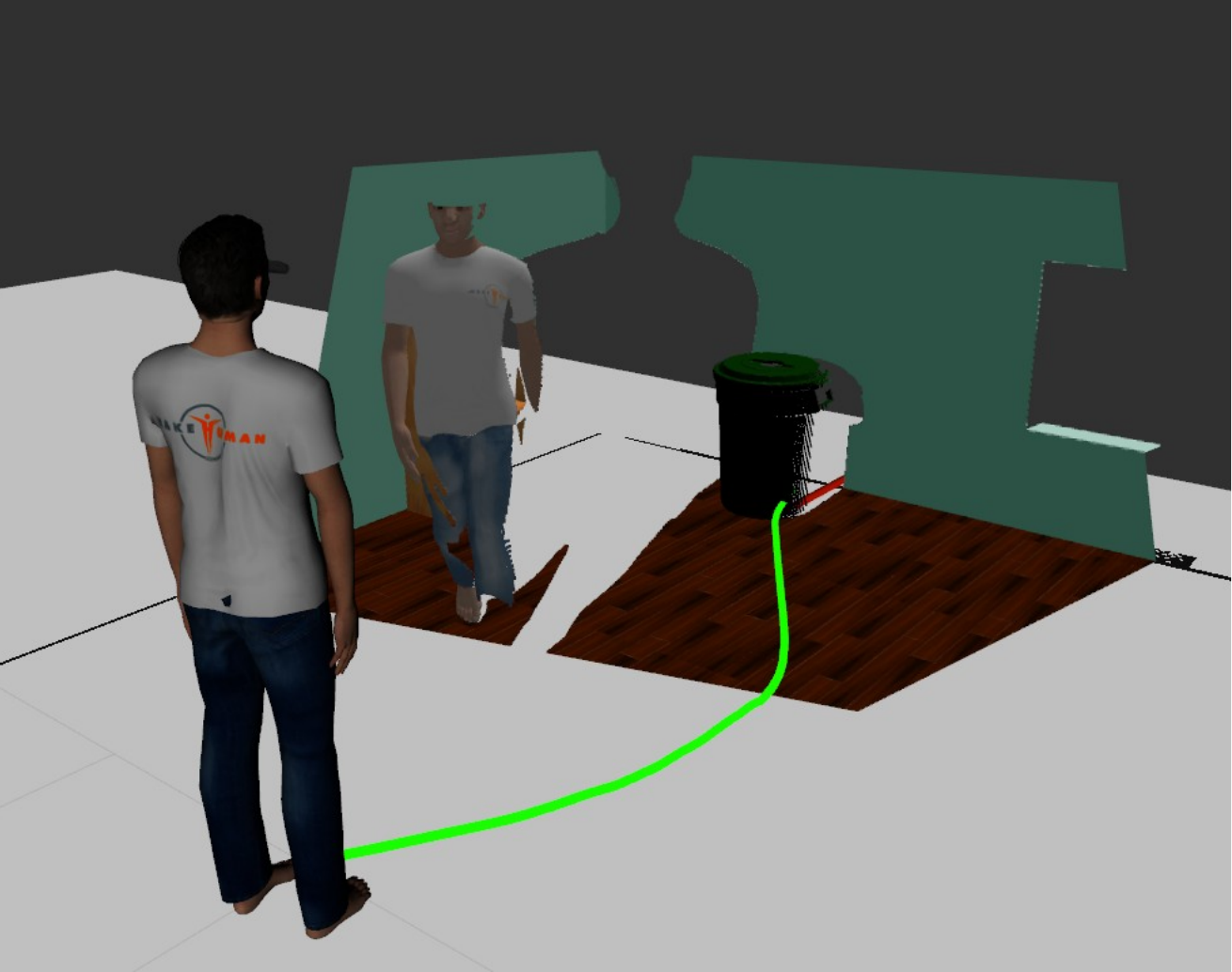}\label{fig4b}}
\subfigure[]{\includegraphics[width = 2in]{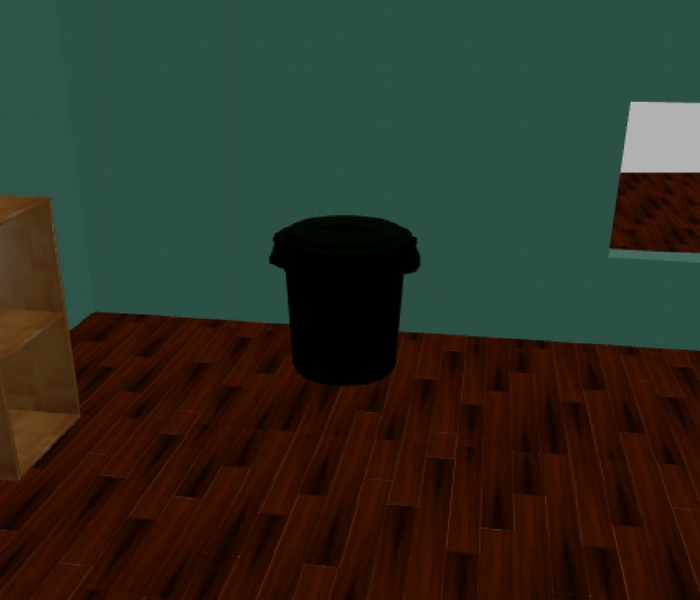}\label{fig4c}}
\subfigure[]{\includegraphics[width=2in]{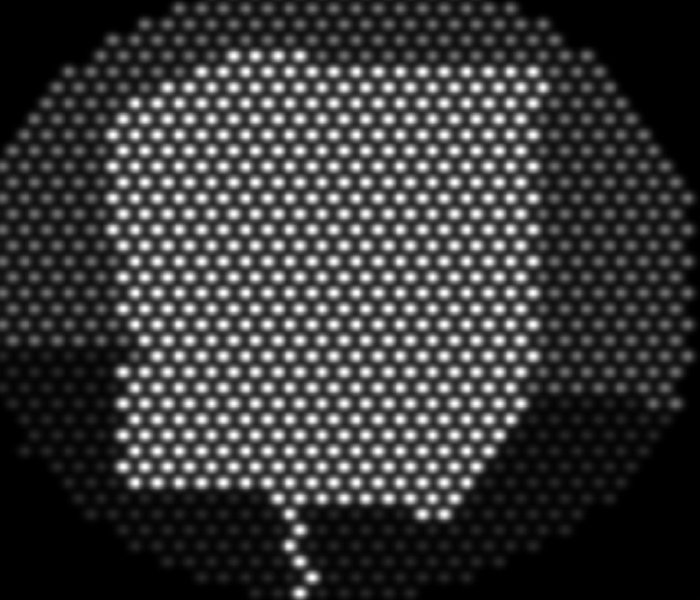}\label{fig4d}}
\subfigure[]{\includegraphics[width = 2in]{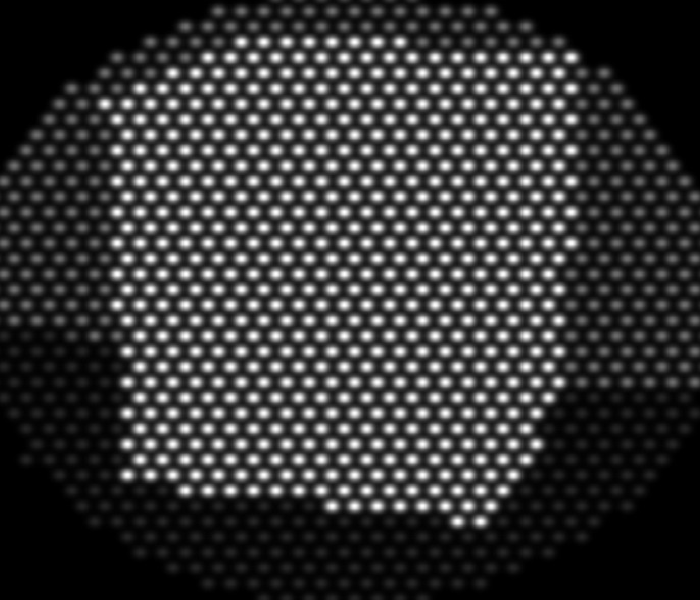}\label{fig4e}}
\subfigure[]{\includegraphics[width = 2in]{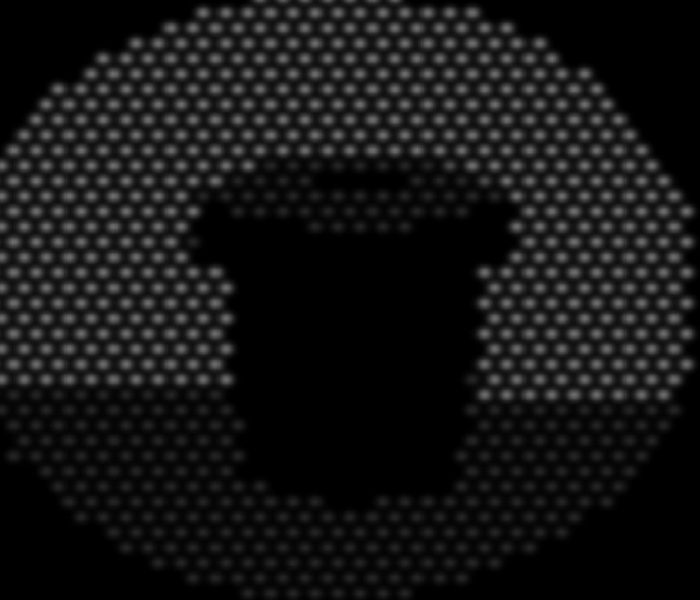}\label{fig4f}}
\caption{\textbf{Guidance methods with bin goal.} In this example, the subject's goal is to find the bin. \subref{fig4a} Simulation environment showing the bin goal. \subref{fig4b} The robotic system obtains an optimized trajectory to the goal avoiding obstacles (green line). \subref{fig4c} Image captured by RGB camera. \subref{fig4d} Image generated by the SPV with the RoboticG method. \subref{fig4e} Image generated by the SPV with the PerceptualG method. \subref{fig4f} Image generated by the SPV with the DirectG method. }
\label{fig4}
\end{figure}

The following subsections describe the three guidance methods used in the experiments. First, our proposal called Robotic Guidance (RoboticG) based on augmented reality navigation system highlights the optimized trajectory or path and the object goals on the phosphene representation. Second, we have used two guidance methods as baselines: a) we remove the path augmentation from our RoboticG method, leaving the goal perception (PerceptualG) and b) we remove both, the path and the goal augmentation from our RoboticG method (DirectG). Examples of the resulting effect on the phosphene images are shown in Figure~\ref{fig3} and Figure~\ref{fig4}.

%%% Figure door Goal

%%% Figure object Goal

% -----------------------------------------------------------
% Robotic Navigation Guidance (RoboticG)
\subsubsection{Robotic Guidance (RoboticG)}

We propose an autonomous navigation system with obstacle avoidance for visual prosthesis called Robotic Guidance (RoboticG). This method performs a path planning that the person has to follow to reach a goal avoiding obstacles and highlight the path with phosphenes (augmented path), as can be seen in Figure~\ref{fig3d} and Figure~\ref{fig4d}. Moreover, we use a goal perception method which segment the goals and highlight them with phosphenes (augmented goal), as can be seen in Figure~\ref{fig3e} and Figure~\ref{fig4e}. We evaluate our RoboticG method using SPV. SPV is made up of Gazebo and simulated phosphenes.

% Diagram 
\begin{figure}[h]
\centering
\includegraphics[width=1\textwidth]{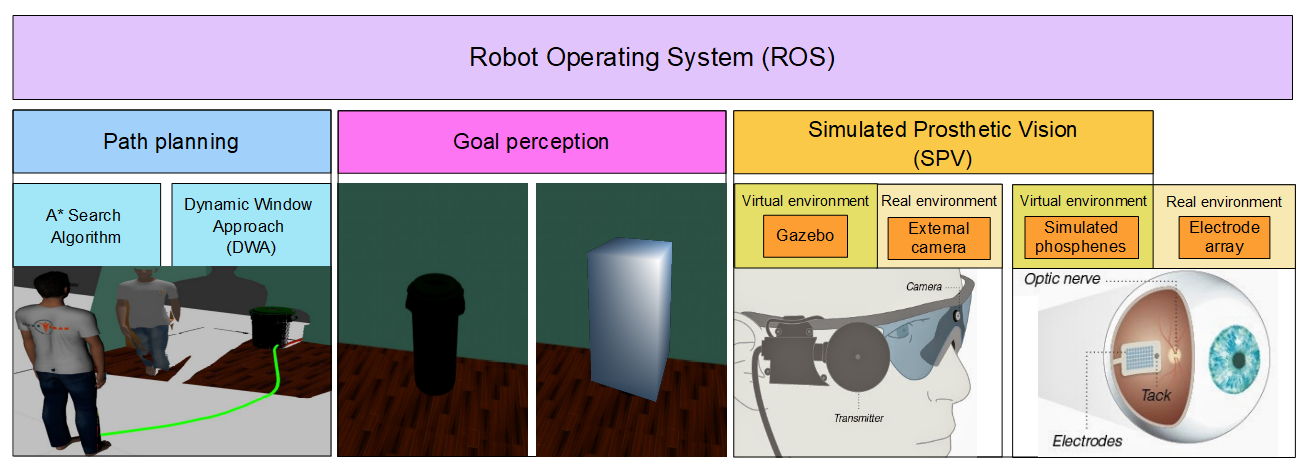}
\caption{\textbf{Diagram of SPV environment.} We use the Robot Operating System (ROS) which handles communications between programs allowing easy control of a robot's mobile operations. We evaluated our RoboticG method composed of Path planning and Goal perception using Simulated Prosthetic Vision (SPV). SPV is made up of Gazebo and simulated phosphenes. One advantage of our method is that in a real environment, Gazebo could easily be exchanged for the external camera of the prosthesis and the simulated phosphenes for the implant electrode array.}
\label{fig5}
\end{figure}

%% Path planning
\paragraph{Path planning}

The most important tasks for navigation are locating the user on the map and the path planning to reach a goal. For the location problem we determine the position and orientation of the subject relative to a 2D map of the virtual environment where it is going to be navigated. One of the great advantages of working in a simulated environment is that it allows us to know the user's location on the map at all times. However, for real experiments, we could use a SLAM system \cite{mur2017orb} which allows locating the subject as long as the scene has previously been mapped. For both, the location system (3D map) and the navigation system (2.5D map), we use a common reference system that allows us to express information from the reference system of one map to the reference system of the other. Our simulation environment is precisely designed so that it can be used with a real sensor. 

% obtained with the depth sensor of the camera

We use two planners for the path planning: a global planner and a local planner. The global planner is based on the A$^*$ Search Algorithm which calculates the optimal path between the source (initial state) and the destination (final state) using the 2D map (see Figure~\ref{fig5}). The local planner is based on the Dynamic Window Approach (DWA) \cite{fox1997dynamic} which following the A$^*$ path, finds a collision-free (`admissible') trajectory considering the obstacles not included in the map. This algorithm uses the information of the obstacles from the 3D point cloud projected on a 2D map, to calculate the most optimal path with which to avoid them. The global trajectory is modified as a function of distance to which obstacles are located, the distance to the target and the alignment of the subject with respect to it. Obstacles are detected in real time.

%% Goal perception
\paragraph{Goal perception}

In order to augment the goals to highlight them, it is necessary to apply some kind of goal perception method, that allow us to recognize the goal element in the scene. Nevertheless, `goal perception' is a broad term and the method to apply may need to be goal-specific. For instance, we could use the Region-Growing algorithm \cite{rabbani2006segmentation}, deep learning detection/segmentation algorithms to identify certain objects in an image \cite{he2017mask}, or advanced 3D perception algorithms that leverage point clouds. Since object detection is not in the scope of this work, we leverage the known map and fix the goals (door and bin) to a certain point in the map so that they are unequivocally identified.

%% Augmented representation with phosphenes
\paragraph{Augmented representation with phosphenes}
 
We use a simple representation with phosphenes for our RoboticG method to represent the direction of the path and the goals. Since in prosthetic vision systems the spatial and photometric resolution is very low, we use a simple representation that is capable of communicating the clues for navigation without obstructing but rather complementing the useful information of the scene. 

Our guidance method consists of representing a line on the floor plane (and projecting it on the image) that follows the path that the person must follow to reach a goal. To represent in the image the sequence of 2D points that leads to a goal calculated by the trajectory planner, we define the location of the goal on the 2D map and project the 3D points onto a phosphene mesh. We assign to these points a high intensity value to clearly differentiate them from the rest of the objects in the scene (augmented path). We also assign a high intensity value to the segmented goals by the goal perception method to differentiate them from the rest of the scene (augmented goal).

\paragraph{Simulated Prosthetic Vision (SPV)}

In this work, we simulated two components of the visual prosthesis: the external camera and the electrode array of the prosthesis, shown in orange in Figure~\ref{fig5}. We simulate the external camera with the virtual environment called Gazebo, which is an open-source 3D robotics simulator \cite{koenig2004design}. The main advantage of Gazebo with respect to other simulators is that it is perfectly integrated with ROS, uses the same type of communication through topics and therefore facilitates a comfortable integration with the rest of the processes. Furthermore, sensor simulations produce messages of the same type as real sensors, so that the same code can be used regardless of the source of the information. One of the main uses of Gazebo is the development of robotics algorithms in controlled simulation environments, so that they can then be applied to real systems without having to modify the algorithm itself. In this way, a person with a visual prosthesis could turn off the implant to stop seeing the real environment and plug in Gazebo seeing the 3D world in the physical implant. One advantage of our method is that in a real environment, Gazebo could easily be exchanged for the external camera of the prosthesis and simulated phosphenes for the implant electrode array. 

In relation to the simulated electrode array, our simulated phosphene map is similar to the frameworks of Sanchez-Garcia et al. \cite{sanchez2020semantic}, McKone et al. \cite{mckone2018caricaturing} and Chen et al. \cite{chen2009simulating}. We approximate the phosphenes as circular dots with a Gaussian luminance profile —each phosphene has maximum intensity at the center and gradually decays to the periphery, following a Gaussian function–. The intensity of a phosphene is directly extracted from the intensity of the same region in the image. For our experiments, each phosphene has eight intensity levels. The size and brightness are directly proportional to the quantified sampled pixel intensities. For our experiment, we use a field of view restricted to $20^\circ$ and 1000 phosphenes. The augmented path and goals are highlighted with a higher phosphene brightness level when the subject is less than 5 m closer from the goal.

\begin{figure}[t]
\centering
\includegraphics[width=0.6\textwidth]{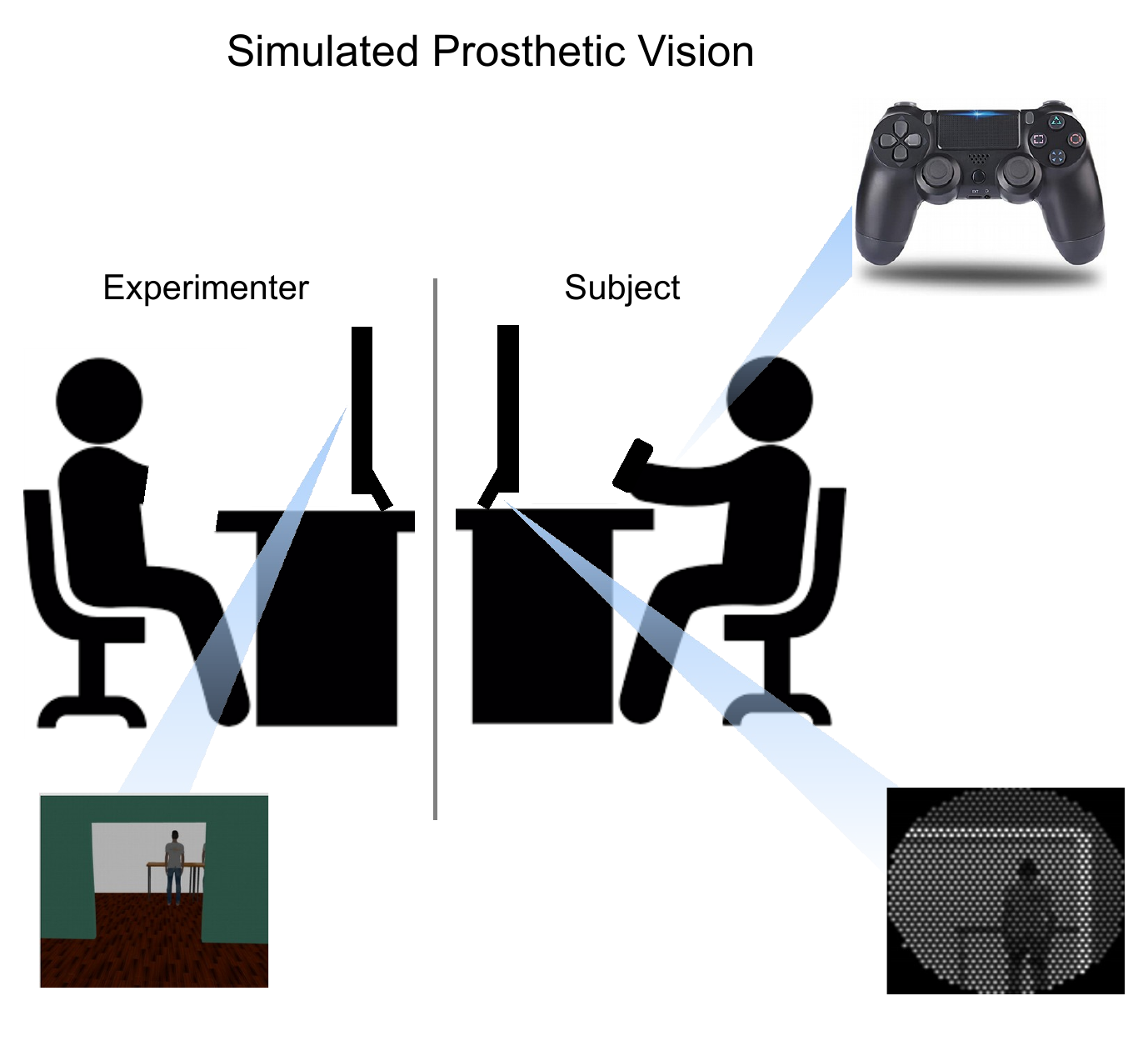}
\caption{\textbf{SPV system.} Subjects were seated on a chair adjusted to their comfort in front of a computer screen holding a video game controller. The experimenter was in front of the subject seeing the same scene but in normal/RGB visualization.}
\label{fig6}
\end{figure}

In this study, we have generated three virtual scenarios with Gazebo and use a dynamic human model in order to simulate the movements performed by a real person. With this two simulators we evaluate the operation of our RoboticG method in the fastest and most comfortable way as a previous step to taking it to a real environment, as can be seen in Figure~\ref{fig6}.

% SPV system

% -----------------------------------------------------------
%% Baseline methods -> PerceptualG and DirectG
\subsubsection{Baseline}

% Ablation is the removal of a component of an AI system
We perform an ablation study as used in AI by removing components from the RoboticG system to determine the significance of each component. First, we remove the path augmentation from our RoboticG method leaving the goal perception, as can be seen in Figure~\ref{fig4e}. This representation is called PerceptualG. Second, we remove any augmentation as can be seen in Figure~\ref{fig4f}. This method is called DirectG. Previous studies have shown that DirectG can be very effective in scenes where high contrast predominates \cite{barnes2012role}.

% -----------------------------------------------------------
% Procedure
\subsection{Procedure}

\begin{figure}[t]
\centering
\includegraphics[width=1\textwidth]{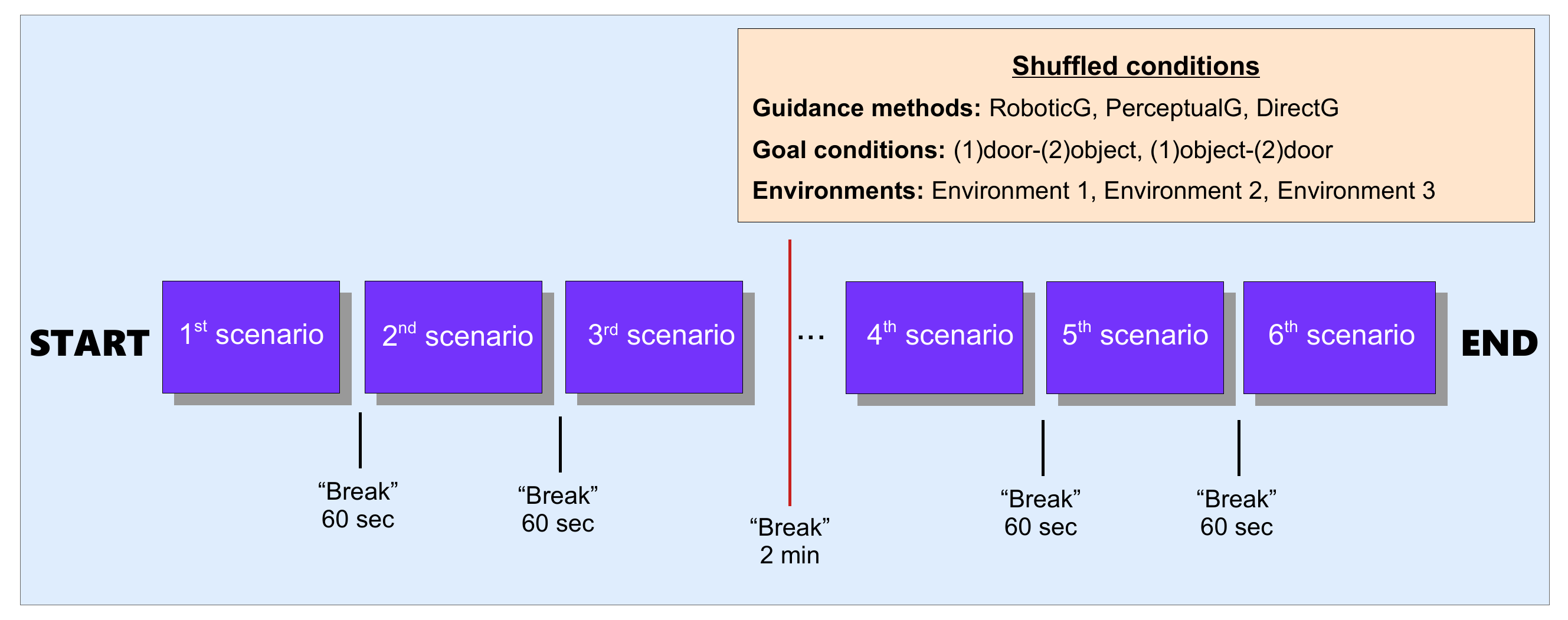}
\caption{\textbf{Trial setup.} To generate the scenario in each step of the trial sequence we used shuffled conditions of environment, goal condition and guidance method. During the experiment, the break time after each scenario was 60 seconds. Break time in the middle of the sequence (between 3th$-$4th scenario) was 2 minutes. The complete experiment took approximately 30 minutes.}
\label{fig7}
\end{figure}

For the formal experiment, participants were recruited to complete a task: finding and reaching a large object while avoiding obstacles in wildly crowded scenarios. Each trial consisted of a sequence of scenarios presented randomly to the subject with the proposed RoboticG method, PerceptualG method and DirectG method, as can be seen in Figure~\ref{fig7}. 

Participants were seated on a chair adjusted to their comfort in front of a computer screen holding a video game controller, as can be seen in Figure~\ref{fig6}. Participants were asked to navigate in a virtual environment while performing various tasks that involved locating an object and crossing the door of the room. Prior to starting the experiment, subjects were shown a training scenario with the normal/RGB visualization, as can be seen in Figure~\ref{fig3c}. The purpose of this training experiment was that subjects could become accustomed with controls, obstacles, goals and the experimental task. Participants were instructed to walk through the environment to familiarize themselves with the video game controller and the simulated environments. Participants also had to demonstrate to the experimenter that they were able to distinguish the goal from the obstacles by walking up to an example of both the door and the object. Following this introductory phase, subjects were given the training scenario with each of the navigation methods (RoboticG, PerceptualG and DirectG) in a random order and they were asked to perform the experimental task of finding the goals without colliding with any of the obstacles. Data from this session was recorded but not included in the analysis. Following the completion of the training trials the experiment was started.

% Trial setup

There were twelve test conditions in total: 3 environments different from training (Environment 1, Environment 2 and Environment 3) (see Figure~\ref{fig2}), 2 goals conditions ((1)door-(2)object: find the door first and then the object; or (1)object-(2)door: find the object first and then the door) and 3 guidance methods (RoboticG, PerceptualG and DirectG), as can be seen in Figure~\ref{fig7}. Participants were instructed to perform six test scenarios with randomly chosen conditions. The participants were asked to walk through the scene until they find the door or object (depending on which goal the experimenter indicated to the subject to find first), avoiding the obstacles in the environment. Session duration was adjusted to allow completion of one trial given the participants’ comfort and including short breaks every scenario. The break time after each scenario was 60 seconds. Break time in the middle of the sequence (between 3th$-$4th scenario) was 2 minutes. The complete experiment took approximately 30 minutes. Data acquisition for each participant was completed after six scenarios. 

We gave additional aid to the participants if they were lost or expressed difficulties, e.g. `take a few steps back' or `turn right'. The number of additional aids was also registered. The performance was assessed according to the time elapsed from the beginning of the experiment until the subject found the second goal as well as the number of collisions. Subject positions were randomized for the three environments and for the goal order condition to decrease potential learning effects.

% -----------------------------------------------------------
% Statistical analysis
\subsection{Statistical analysis}

Data were analyzed using two-tailed paired t-test with Tukey’s correction to evaluate simultaneously the effect of the navigation methods (RoboticG, PerceptualG and DirectG) on the response variables covered distance and navigation time with $p=$ 0.05, * $<$ 0.05 , ** $<$ 0.01, *** $<$ 0.001 and \emph{ns} not significant.

% -----------------------------------------------------------
% Results
\section{Results}

\begin{figure}[t!]
\centering
\subfigure[]{\includegraphics[width = 3in]{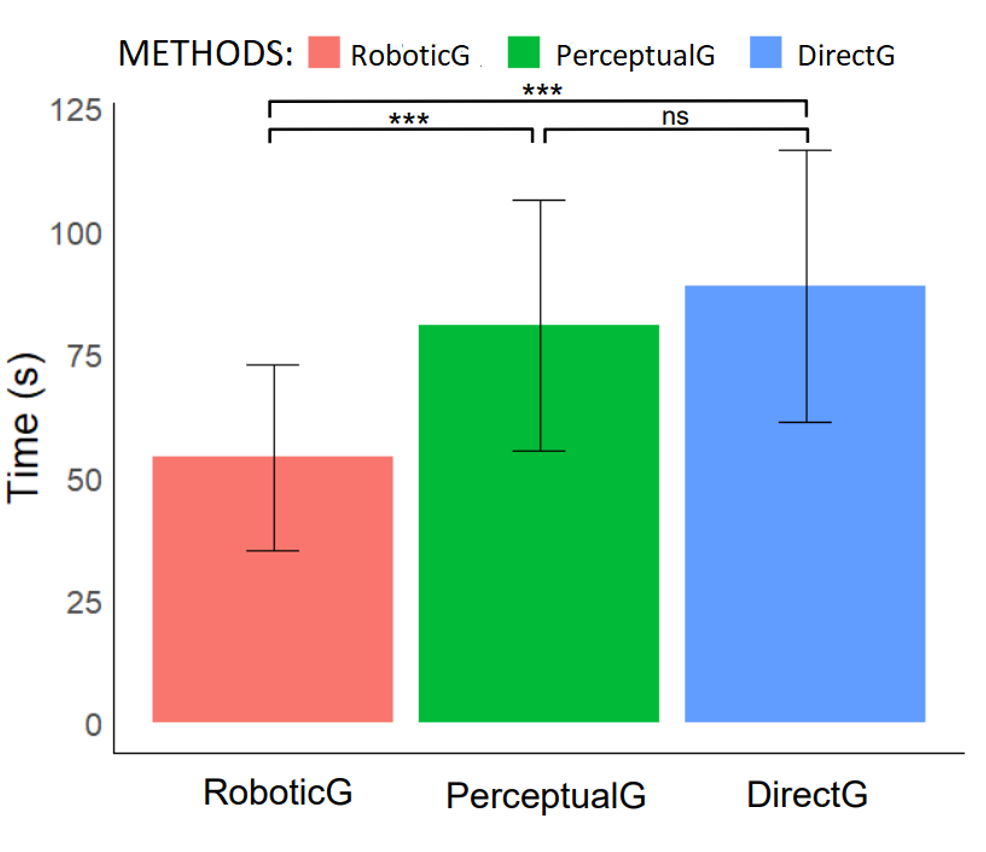}\label{fig8a}}
\subfigure[]{\includegraphics[width = 3in]{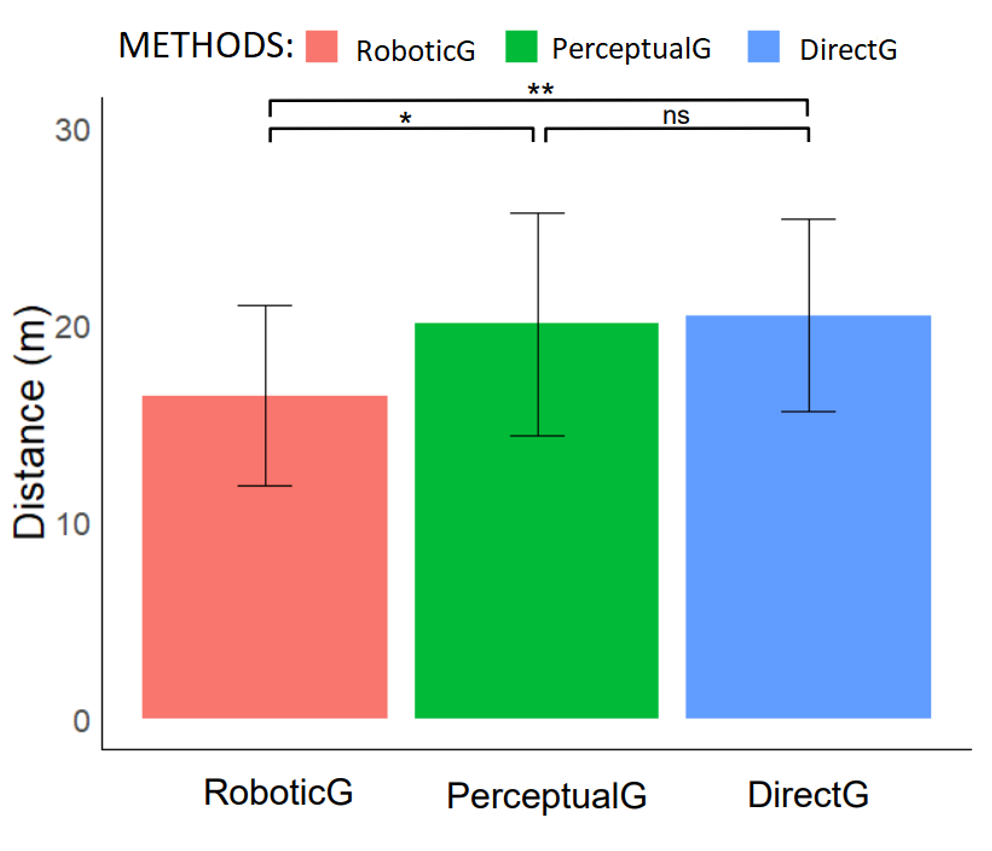}\label{fig8b}}
\caption{Mean time and distance required to reach both goals for the three guidance methods. \subref{fig8a} Time results for \textit{RoboticG}, \textit{PerceptualG} and \textit{DirectG} methods. High scores indicate that the subjects needed more time to perform the location and avoiding obstacles task. \subref{fig8b} Distance results for \textit{RoboticG}, \textit{PerceptualG} and \textit{DirectG} methods. High scores indicate that subjects covered a longer trajectory to reach the goals.  ***=p$<$.001; **=p$<$.01; *=p$<$.05; ns=p$>$.05. All t-tests paired samples, two-tailed. }
\label{fig8}
\end{figure}

The performance for the three guidance methods is summarized in Figure~\ref{fig8}. The results show the time and the distance (mean $\pm$ standard deviation) for aggregated data from all subjects and all environments. Time (in seconds) is the time from the subject’s first step until the subject reach the goal. The distance (in meters) is defined as the covered distance since the subject start walking until the subject reach the goal. We also performed a test to determine if the mean difference between specific conditions are statically significant using two-tailed test with a significance level $\alpha=0.05$. The number of obstacle contacts or bumps within a trial was recorded as the number of times the distance between the center of an obstacle and the center of the subject’s body is less than 1 m. The total number of bumps per method is shown in Table~\ref{tab:table1}.

Figure~\ref{fig8a} shows the time performance for the three guidance methods (RoboticG, PerceptualG and DirectG). For the DirectG method, the average performance is 88.80 $\pm$ $27.55$. For the PerceptualG the average performance is 80.91 $\pm$ $25.47$. For the RoboticG the average performance is 54.02 $\pm$ $18.87$. No significance difference were found for the DirectG and PerceptualG method ($p=0.2512$), as time performance was very similar for both method. However, very significant differences were found for the RoboticG-PerceptualG and RoboticG-DirectG.

% Table 1
\begin{table}
\caption{\label{tab:table1} Mean value of time, distance and total number of bumps for the three methods (RoboticG, PerceptualG and DirectG). $*** = p<.001$; $** = p<.01$; $* = p<.05$; $ns = p>.05$.}
\begin{indented}
\item[]\begin{tabular}{ccccc}
\br
Guidance &  Time & Distance  & Total & Aids / \\
  methods & (s) & (m) & bumps & Interventions \\
\mr
RoboticG &  54.02 $\pm$ $18.87$ & 16.42 $\pm$ $4.59$ & 4 & 0 \\
PerceptualG & 80.91 $\pm$ $25.47$ & 20.03 $\pm$ $5.67$ & 27 & 6 \\
DirectG &  88.80 $\pm$ $27.55$ & 20.48 $\pm$ $4.85$ & 24 & 9 \\
\br
\end{tabular}
\end{indented}
\end{table}
\normalsize

Figure~\ref{fig8b} shows the covered distance for the three guidance methods. For the DirectG method, the average performance is 20.48 $\pm$ $4.85$. For the PerceptualG the average performance is 20.03 $\pm$ $5.67$. For the RoboticG the average performance is 16.42 $\pm$ $4.59$. No significance difference were found for the DirectG and PerceptualG method ($p=0.7812$), as distance performance was very similar for both method.

Table~\ref{tab:table1} shows the total number of bumps for all the guidance methods. The PerceptualG method has more number of bumps than other methods. Interestingly, the methods DirectG and PerceptualG have almost the same number of collisions. However, it makes sense because in the PerceptualG method does not locate the goals until the subject is not close to them. This means that until reaching the goals the PerceptualG method is basically the same as the DirectG method.

Figure~\ref{fig8} shows box-plots of data distribution with 25, 50 and 75\textit{th} quartiles for time and distance. In the case of required time to reach both goals (see Figure~\ref{fig9a}), there is difference between DirectG-RoboticG and PerceptualG-RoboticG. There is also difference between DirectG-PerceptualG. In Figure~\ref{fig9b}, the medians of the DirectG and PerceptualG methods are similar. However, comparing the medians of the DirectG and PerceptualG methods with RoboticG they are well separated, being the median for the RoboticG around 15m.

% https://jov.arvojournals.org/article.aspx?articleid=2193740

% -----------------------------------------------------------
% Discussion
\section{Discussion}

Efficient navigation is a fundamental capability of every person to have autonomy on our daily life. People have their own well-defined criteria for choosing a path when moving from one place to another. But visually impaired people are not privileged to make their own choices when navigating. Significant amount of work has already been done for developing navigation systems for visually impaired people so that they could independently reach a target, without any kind of external assistance. Some examples of these systems are ultrasonic sensors, Global Positioning System (GPS) or radio-frequency identification \cite{joseph2014intelligent,kanwal2015navigation}. Very few researchers have focused on developing routing algorithms for the visually impaired \cite{nandini2019novel,kammoun2010route,swobodzinski2009indoor}.

% Box-plots
\begin{figure}[t!]
\centering
\subfigure[]{\includegraphics[width = 3in]{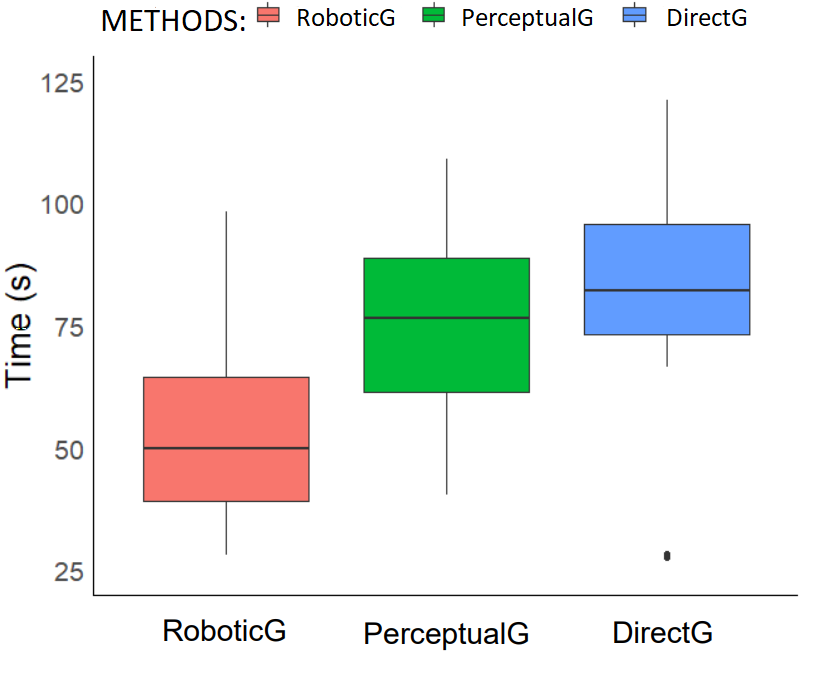}\label{fig9a}}
\subfigure[]{\includegraphics[width = 3in]{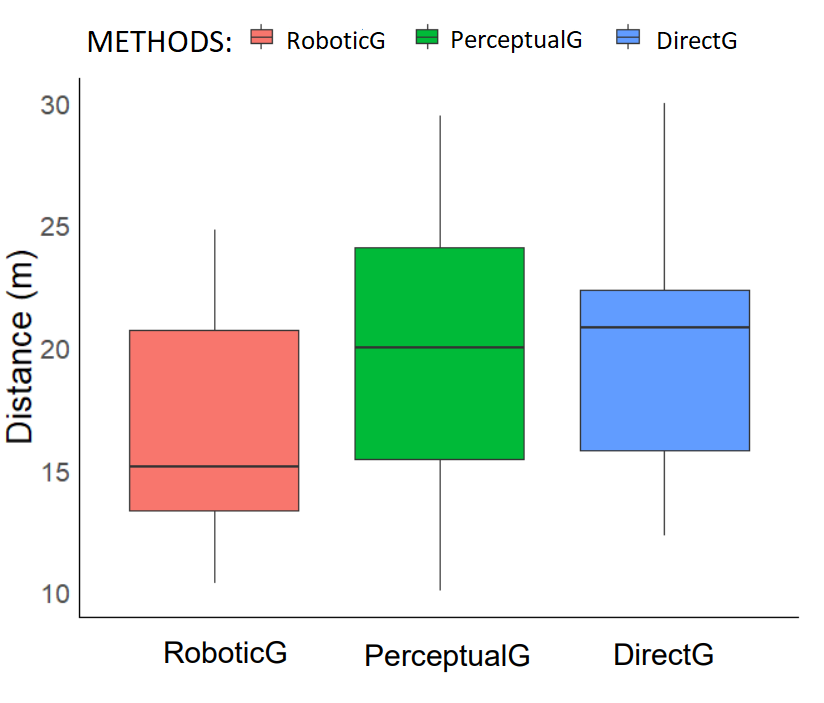}\label{fig9b}}
\caption{Mean time and distance required to reach both goals for the three guidance methods. \subref{fig9a} Box-plot for time. \subref{fig9b} Box-plot for distance. }
\label{fig9}
\end{figure}

In this work, we propose an augmented navigation system inspired on robotics research, called RoboticG, which combines a path planning algorithm and an obstacle avoidance method for routing the visually impaired person through an obstacle free optimal path. There have been various methods introduced by researchers in different context to study path planning problem \cite{lavalle2001randomized,hart1968formal,kavraki1996probabilistic}. In the field of robotic, path planning is usually the first step to navigate a robot, where the final step incorporate techniques for obstacle avoidance \cite{minguez2005sensor,khatib1986real,barraquand1997random}. %Other investigators used a technique called Grid based path planning which use metaheuristic techniques such as genetic algorithm \cite{alajlan2013global}, harmony search \cite{yang2009harmony} and quad harmony search \cite{koceski2014novel}.

We evaluated our RoboticG system using simulated prosthetic vision (SPV) composed by Gazebo \cite{koenig2004design}, a virtual environment simulator, and the Robotic Operating System (ROS) \cite{quigley2009ros}. Virtual environments can be built to implement the intended experimental platform without constraints associated with the real world. Furthermore, the use of virtual environments can help to avoid the experimental error caused by collisions and the use of touch to identify objects indirectly. Virtual environments have been widely used for experimentation in navigation tasks \cite{zhao2017recognition,vergnieux2012spatial,dagnelie2007real}. Some SPV research such as Vergnieux et al. \cite{vergnieux2012spatial} used a virtual indoor environment to investigate the navigation capabilities that could be restored through two different stimulation strategies consisting in a reduction of the environment view to match the number of electrodes and an object recognition algorithm in order to present only recognized elements. Wang et al. \cite{wang2008virtual} created a virtual maze with SPV to assess navigation performance under contrasting conditions of varying luminance, background noise, and phosphene loss.

We found that object localization is well achieved with low resolution and restricted field of view using our guidance method (RoboticG). In a study by Golledge et al. \cite{golledge1995path}, they conducted on the criteria of selection of roads or routes of an average human being and showed that the best classified criteria are the shortest distance, the least time and the least number of turns. As can be seen in Figure~\ref{fig8a}, we obtained a significant reduction in the navigation time until reaching the goals, compared to other baseline methods which do not used a navigation system, the PerceptualG and DirectG methods. Besides, participants took a shorter path with the RoboticG method compared with the baseline methods (see Figure~\ref{fig9}). This is intuitive since the navigation algorithm RoboticG calculates the shortest path to reach the goals. Contrary, the participants took longer to reach the goals without the navigation system. Further, no significant difference was obtained with the baseline methods. This makes sense since the PerceptualG method locates the goals when the subject is at a distance less than 5m. From a real implementation, using the A* search algorithm we need to know where the goal is, but in practice the goal can be moved. Therefore, our navigation system gives an estimated 2D location of the goals. But when it comes to highlighting the goal we need a precise 3D location. In our case, the bin is placed on the ground but if the bin was not on the ground, we would need delicate perception. Therefore, our method leads the subject to an approximate location and identifies targets at 5 m. Until the subject is close enough to the targets, both the PerceptualG and DirectG methods are similar.

Obstacle avoidance also becomes a crucial point when dealing with visually impaired people. In any day-to-day environment, a visually impaired person may encounter various types of obstacles, whether dynamic or static. Early prototypes of obstacle detection devices were based on radar and sonar systems \cite{national1986working,benjamin1974laser,armstrong1973summary}. Recently, some of the available assistance systems focus on ringing a bell when an obstacle is detected on the road. However, these systems are unable to redirect the user from their current position to their desired destination once an obstacle has been detected. Our results show that the RoboticG method reduce drastically the number of bumps during the navigation due to the recalculation of the trajectory once an obstacle has been detected (see Table~\ref{tab:table1}). One of the objectives of navigation systems is to calculate the optimal route that the user can take from his initial position to the destination. Figure~\ref{fig9} shows examples of path trajectories for the three guidance methods. In the three environments the path trajectories from the initial position to the goals were shorter and smoother with the RoboticG method. The PerceptualG and DirectG methods do not show much difference between them in environments 2 and 3. Although when we analyze it in a more complicated environment (with more obstacles) such as environment 1, we can see how an object recognize algorithm, PerceptualG, achieve a more direct path than the DirectG method which presents more oscillations. Moreover, our results suggest that subjects do not need additional aid with the navigation system from another person, in this case the experimenter (see Table~\ref{tab:table1}). Although the RoboticG method has shown good results compared to the baseline methods, we believe that using only the augmented path of the RoboticG method (removing the augmented goal) would achieve very similar results.

% Trajectory
\begin{figure}[t!]
\centering
\subfigure[]{\includegraphics[width = 2in]{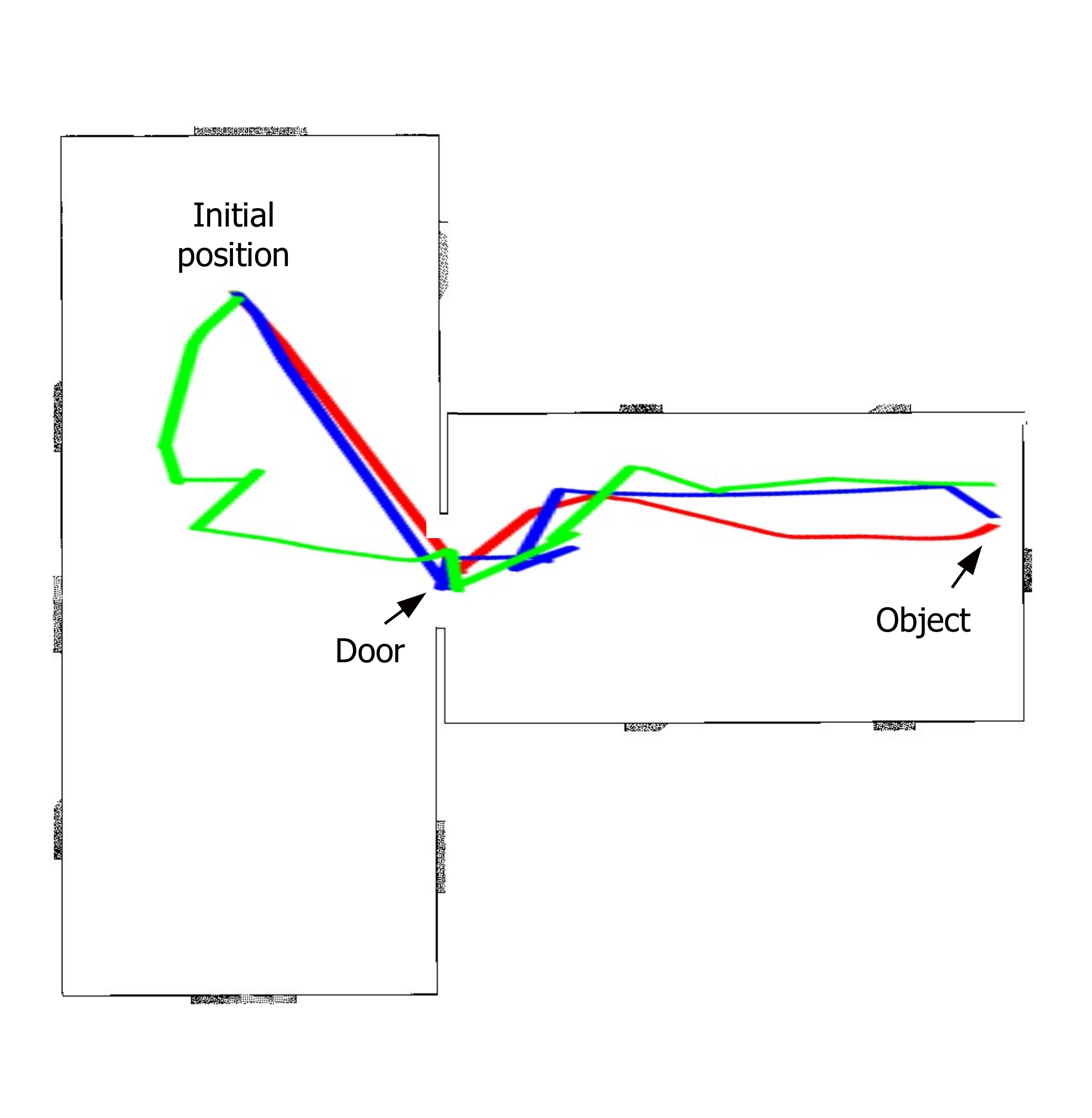}\label{fig10a}}
\subfigure[]{\includegraphics[width = 2in]{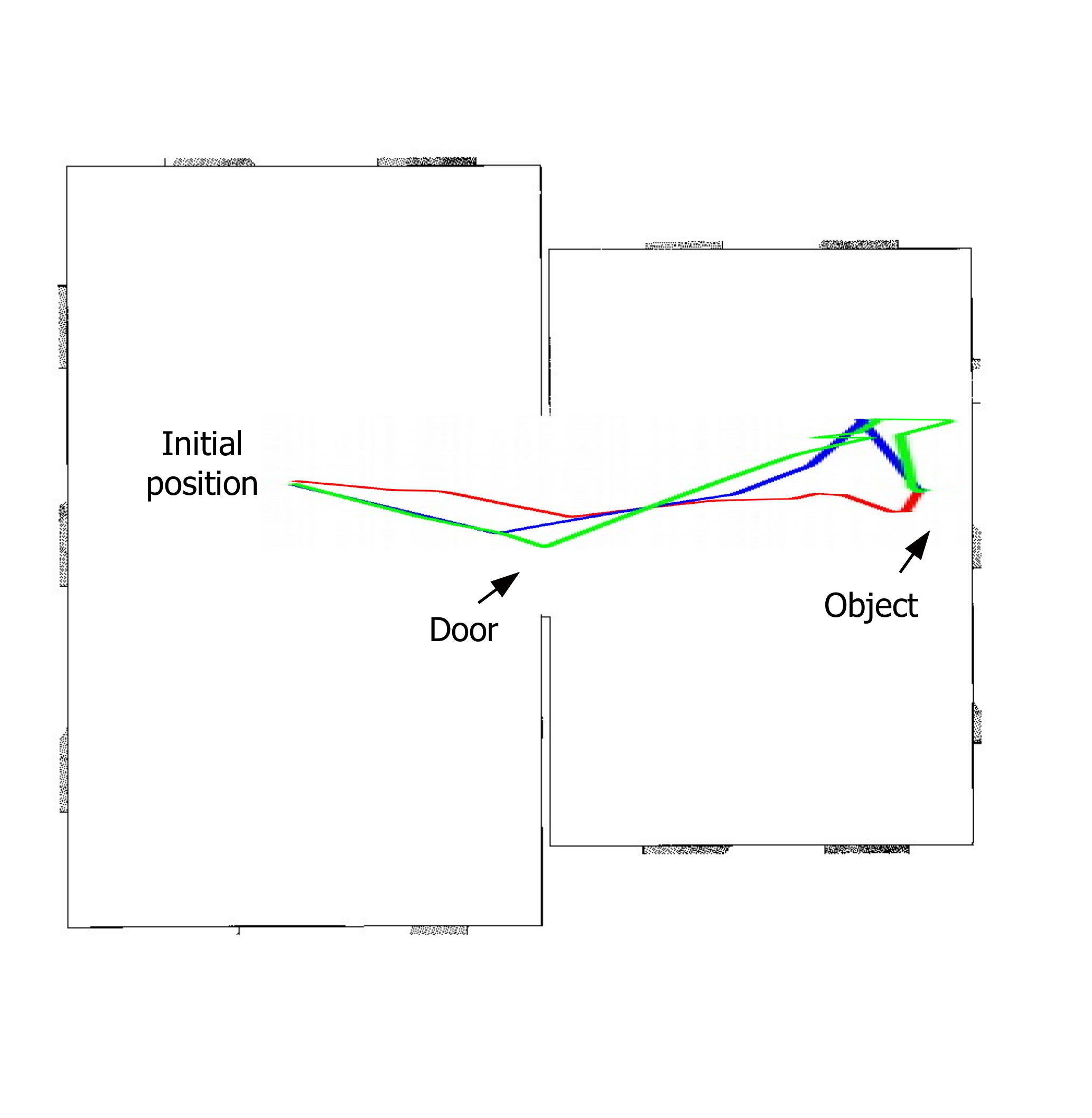}\label{fig10b}}
\subfigure[]{\includegraphics[width = 2in]{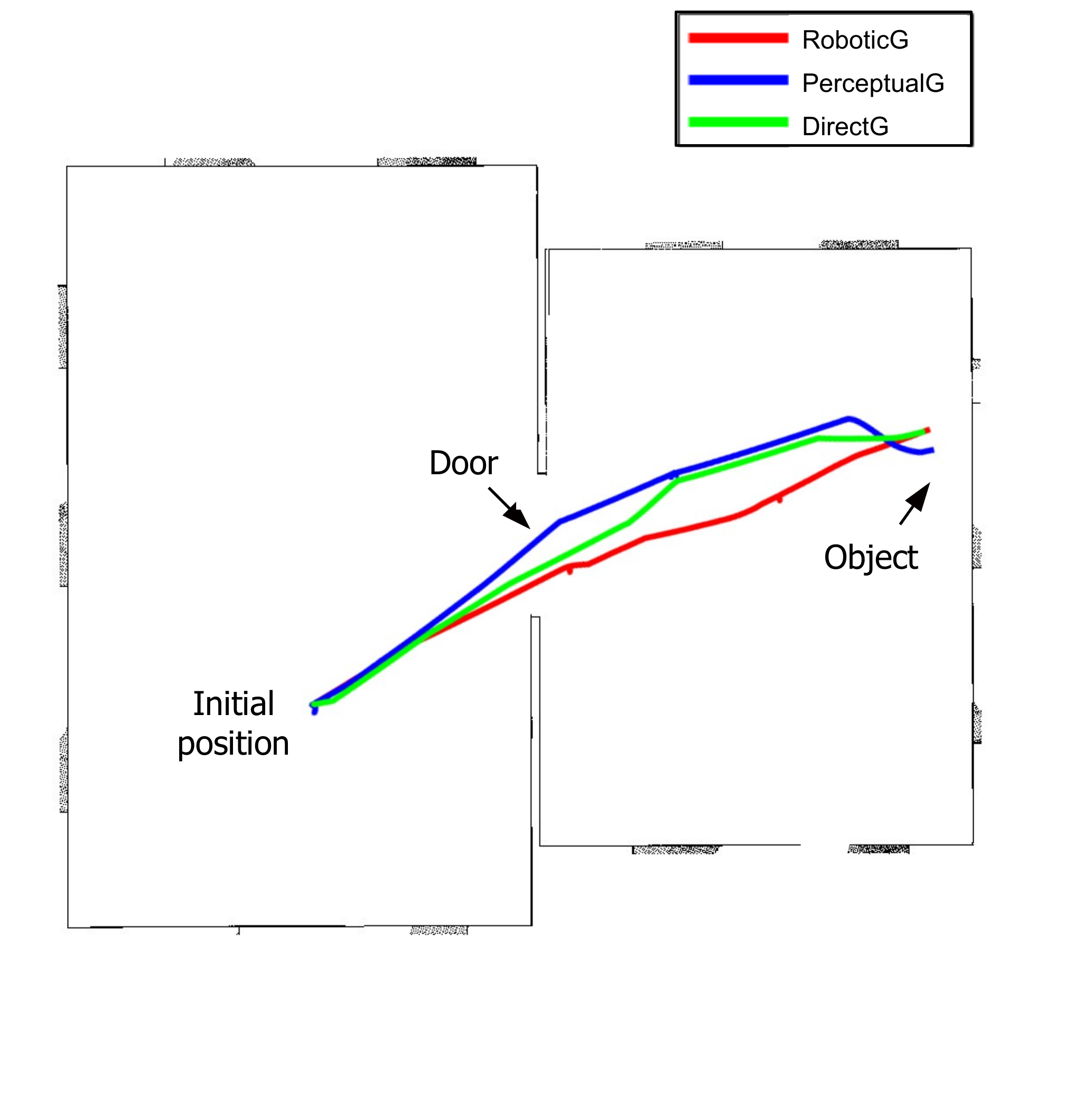}\label{fig10c}}
\caption{Examples of path trajectories for the three guidance methods (RoboticG, PercetualG and DirectG), during the experiment. \subref{fig10a} Path trajectories in Environment 1. \subref{fig10b} Path trajectories in Environment 2. \subref{fig10c} Path trajectories in Environment 3.}
\label{fig10}
\end{figure}

This system has numerous tasks that have to be working simultaneously: point cloud processing, location on a map, path planning and generation of the phosphene image with the trajectory. All this is a challenge at the implementation level, since the numerous software packages involved have to communicate with each other and send the necessary information to each other on time. Our RoboticG implementation allows working in a simulation environment, which allows repeatability in the design of experiments with real subjects and gives statistical meaning to the results. Furthermore, the proposed guidance system works on ROS. ROS has been used to communicate and program the different systems of this project, a set of libraries and tools developed mostly in C ++ focused on the development of robotic applications. One of the advantages of using a robotics framework such as ROS is that it allows us to extend the initial approach so that the system can be used in real scenarios using an RGB-D camera or using a SLAM system for localization \cite{mur2017orb}.

% -----------------------------------------------------------
% Conclusion
\section{Conclusions}

Navigation in day-to-day environments is fundamental for all humans. People with visual implants still require constant assistance for navigating from one location to another. Hence there is a need for a system that is able to assist them safely during their journey. We propose an augmented navigation system with obstacle avoidance for guidance in visual prosthesis which points out the path that the subject has to follow to reach the goals. In our study participants navigated through a virtual environment which allowed us to systematically study navigation performance based on prosthetic vision. The task consisted on object localization and obstacle avoidance during navigation. Our results show that our RoboticG system help navigation performance by reducing the navigation time and walking distance to reach the goals, even significantly reducing the number of obstacles collisions, compared to other baseline methods.

% Acknowledgments
\section*{Acknowledgments}

This work was supported by project RTI2018-096903-B-I00 (MINECO/FEDER, UE) and BES-2016-078426 (MINECO). The authors thank Lorenzo Montano Oliván and Jesús Bermúdez Cameo for collaborating in the navigation system development.

\section*{References}

\bibliographystyle{iopart-num}
\bibliography{master}

\end{document}